\pdfoutput=1
\documentclass[11pt]{article}
\usepackage{ACL2023}

\usepackage{times}
\usepackage{latexsym}

\usepackage{times}
\usepackage{latexsym}
\usepackage{amsmath}
\usepackage{amssymb}
\usepackage{bm}
\usepackage{todonotes}
\usepackage{algorithm}
\usepackage{algorithmicx}  
\usepackage{algpseudocode}  
\usepackage{color}
\usepackage{cleveref}
\usepackage{booktabs}
\usepackage{multirow}
\usepackage{stfloats}
\usepackage{pifont}
\usepackage{xcolor}
\usepackage{adjustbox}
\usepackage{setspace}
\usepackage{diagbox}
\crefname{section}{§}{§§}
\Crefname{section}{§}{§§}

\usepackage[T1]{fontenc}

\usepackage[utf8]{inputenc}

\usepackage{microtype}
\usepackage{inconsolata}

\usepackage{graphicx}

\title{Adaptive and Personalized Exercise Generation\\ for Online Language Learning}

\author{
Peng Cui~\qquad~Mrinmaya Sachan \\
Department of Computer Science, ETH Zürich \\
\texttt{
\{\href{mailto:peng.cui@inf.ethz.ch}{peng.cui}, \href{mailto:mrinmaya.sachan@inf.ethz.ch}{mrinmaya.sachan}\}@inf.ethz.ch 
}\\
}

\begin{document}
\maketitle
\begin{abstract}
Adaptive learning aims to provide customized educational activities (e.g., exercises) to address individual learning needs. 
However, manual construction and delivery of such activities is a laborious process. 
Thus, in this paper, we study a novel task of \emph{adaptive} and \emph{personalized} exercise generation for online language learning.
To this end, we combine a knowledge tracing model that estimates each student's evolving knowledge states from their learning history and a controlled text generation model that generates exercise sentences based on the student's current estimated knowledge state and instructor requirements of desired properties (e.g., domain knowledge and difficulty).
We train and evaluate our model on real-world learner interaction data from Duolingo and demonstrate that LMs guided by student states can generate superior exercises.
Then, we discuss the potential use of our model in educational applications using various simulations. 
These simulations show that our model can adapt to students' individual abilities and can facilitate their learning efficiency by personalizing learning sequences.\footnote{Our implementation is available at \href{https://github.com/nlpcui/AdaptiveQG}{https://github.com/ nlpcui/AdaptiveQG.}}

\end{abstract}

\section{Introduction}
\begin{figure}[ht]
\centering
\includegraphics[width=\columnwidth]{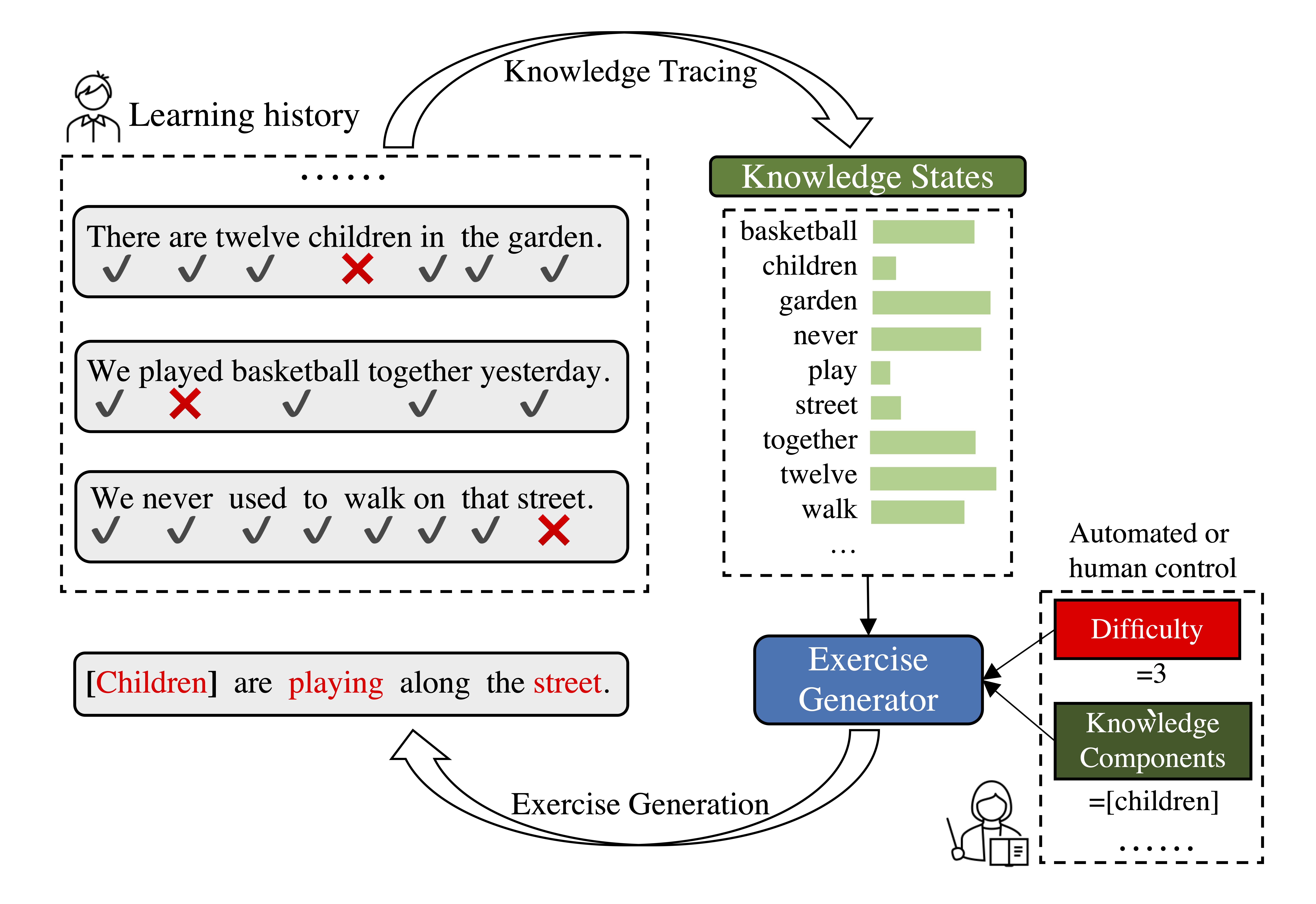}
\caption{
We first assess student knowledge states from their learning history and then generate exercises based on estimated states and instructor control of desired properties including domain knowledge (vocabulary) and difficulty levels (expected error numbers).
}\label{fig:task_illustration}
\end{figure}
Adaptive learning technologies which continuously monitor student progress to  dynamically adjust the level or type of learning materials based on the individual’s abilities are quite popular \citep{becker2018horizon}. 
Empirical studies have shown various benefits of adaptive learning, such as improved student learning outcomes \citep{bailey2018making,holthaus2019recommendation}, lower dropout rates \citep{daines2016improving}, and increased instructor satisfaction \citep{lessonsFrom}. 
Despite their effectiveness, designing adaptive systems is challenging as it usually involves planning a series of exercises that is personalized and adaptive to each student, which requires diverse exercise planning as well as an understanding of the student learning process.

On the other hand, powered by advances in neural NLP, works have been done for automatically generating text-based exercises or questions for educational purposes in second language learning \citep{heck2022parametrizable,perez2017multilingual}, mathematics \citep{polozov2015personalized,zhou-huang-2019-towards,wang-etal-2021-math}, and computer science \citep{susanti2017controlling}. 
Nevertheless, how to apply these approaches in adaptive systems remains an open question.
First, existing methods largely rely on pre-defined question templates or specified information sources (e.g., a passage), thereby resulting in limited knowledge coverage and low question difficulty control, and as a consequence, do not meet each student's individual and nuanced learning needs.
Besides, they are usually designed to generate standalone exercises, whereas adaptive learning systems usually require a continuous supply of exercises.
Another related line of research studies exercise recommendation to customize learning content based on individual capabilities and goals \citep{wu2020exercise,huang2022design}.  
However, these systems are limited by the diversity of the exercise pool.

To address the above limitations, we study the task of exercise generation in the context of adaptive learning, where we hypothesize that a student's \emph{dynamic knowledge state} holds the key to generating \emph{adaptive} and \emph{personalized} exercises. 
Specifically, we ground our study in the domain of language learning to create exercise sentences for translation, of which Figure \ref{fig:task_illustration} illustrates the overall process. 
We start with an assumption about the dynamics between exercise difficulty, vocabulary, and a student's knowledge state (\cref{3}).
Then, we propose an approach (\cref{4}) that marries knowledge tracing (KT; \citet{corbett1994knowledge}), a technique for estimating students' mastery states of knowledge components from their learning history, with a controlled text generation model that generates the next exercise based on instructor requirements, such as specified \emph{domain knowledge} and \emph{target difficulty}. 
We further explore various strategies to adapt the generation of exercises based on students' changing knowledge states.
In doing this, our model not only supports personalized generation where the instructor (or the system) can express some desired properties of the generated exercises but is also adaptive to each student's learning progress. 

We conduct extensive experiments on real-world student learning data from Duolingo\footnote{\href{https://www.duolingo.com/}{https://www.duolingo.com/}}, a popular online language learning platform that offers structured and individualized learning content.
Our results (\cref{6}) show that pre-trained LMs can help KT assess student language knowledge while student states estimated by KT can guide LMs to generate adaptive and personalized exercises.
We further discuss the potential use of our model in educational applications with simulations. 
The simulations show that our model can dynamically adjust exercise difficulty to match individual learning progress and facilitate their learning efficiency by customizing exercise sequences.

\section{Related Work}
\textbf{Adaptive Learning} technologies that dynamically monitor student progress and adjust the course content based on an individual’s abilities have demonstrated various benefits in education \citep{becker2018horizon}.
Such systems usually consist of three core components: (1) a \emph{domain model} which refers to the content and structure of the topic to be taught, (2) a \emph{learner model} which repeatedly measures and updates learner characteristics, and (3) an \emph{adaption model} which combines information from the domain and learner model to offer adaptive instructions \citep{vagale2012learner,imhof2020implementation}. In this study, we build the learner model based on the KT technique and combine the domain and adaption model into an LM which generates learning content adaptively based on user features captured by the learner model.

\noindent \textbf{Knowledge Tracing} \citep{corbett1994knowledge} is the technique to estimate students' knowledge mastery ${\bf s}$ from their practiced exercises (${\bf e}$) and responses (${\bf r}$): 
\begin{equation}
    \begin{aligned}
        {\bf s_{t+1}} = f_{KT}(({\rm e_{1}}, {\rm r_{1})}, ({\rm e_{2}}, {\rm r_{2}}), ..., ({\rm e_{t}}, {\rm r_{t}})).
    \end{aligned}
\end{equation}
Early KT approaches model $f_{KT}$ as variants of logistic regression, such as Item Response Theory (IRT) and Additive Factor Model (AFM) \citep{cen2008comparing}, or probabilistic models such as Bayesian Knowledge Tracing \citep{corbett1994knowledge} and its variants \citep{yudelson2013individualized,kaser2017dynamic}. 
These approaches heavily rely on their assumptions of the learning process which are often incomplete. 
In recent years, neural networks have become the dominant method in this area. 
\citet{piech2015deep} proposed the first Deep Knowledge Tracing model based on Recurrent Neural Networks. 
After that, various architectures have been applied to model different characteristics of learning, such as self-attention \citep{PandeyK19-0,shin2021saint+}, memory networks \citep{abdelrahman2019knowledge}, and graph neural networks \citep{tong2020structure}.

\noindent \textbf{Exercise Generation}.
Previous exercise generation approaches for language learning primarily retrieve and manipulate text to create fixed types of exercises, such as gap fill and multiple-choice exercises \citep{agarwal2011automatic,perez2017multilingual,heck2022parametrizable}, which are limited by the richness of the corpus. 
Besides them, some Question Generation (QG) approaches have been proposed for educational purposes \citep{zhao2022educational,wang-etal-2021-math}. 
While some of them allow for user control of certain question properties, they do not consider learners' individual and dynamic learning needs and progress. Thus, they cannot achieve the goal of adaptive learning.   
Recently, \citet{srivastava-goodman-2021-question} proposed an adaptive question generation model that connects question difficulty with student knowledge.
However, it neither models students' fine-grained knowledge states nor provides control over domain knowledge.
Consequently, it is insufficient for practical use.

\noindent \textbf{Controlled Text Generation (CTG)} methods aim to steer text generation toward certain attributes.
Existing CTG approaches can be broadly classified into three types: directly training a class-conditional language model (CCLM) \citep{keskar2019ctrl,ziegler2019fine,ficler-goldberg-2017-controlling}, guiding a 
model via an attribute discriminator \citep{Dathathri2020Plug,liu-etal-2020-data}, or manipulating decoder's logits (also referred to as weighted decoding) \citep{holtzman-etal-2018-learning,yang-klein-2021-fudge}.
This study explores difficulty and lexical control in generating language learning exercises.
Additionally, we seek to adapt the model's controllability to different users by building the dependency between control signals and individual states. 

\section{Problem Formalization}\label{3}
Let $\mathcal{H}_{\leq n}=\{(e_1, r_1), ..., (e_n, r_n)\}$ be a student's {\bf learning history} consisting of $n$ exercises and responses. Here, $e_{i}=\{w_{i,1}, ..., w_{i,|e_{i}|}\}$ is an {\bf exercise sentence} for translation and $r_i \in \{0,1\}^{|e_{i}|}$ is the {\bf correctness label} for each word in $e_{i}$. We generate the next exercise $e_{n+1}$ based on: 
\begin{itemize}
    \item $C_{n+1}$: {\bf knowledge components} that should be involved in $e_{n+1}$. In language learning, we consider a word as a knowledge component, and therefore $C_{n+1}=\{c_{1}, ..., c_{|C_{n+1}|} | c_{*}\in \mathcal{V} \}$ is a subset of vocabulary $\mathcal{V}$ that should be included in the output. In general, the knowledge components can be user or system defined based on the current learning material.
    \item ${\bf s_{n+1}}$: a student's {\bf knowledge state} for the knowledge components (the vocabulary) after $n$ interactions. ${\bf s_{n+1}}$ can be formalized as a $|\mathcal{V}|$-dimensional vector with each entry between 0 and 1 indicating the mastery probability of that word. 
    \item $d_{n+1}$: the {\bf expected difficulty} of $e_{n+1}$. We use individual performance to estimate problem difficulty. For a particular student, the difficulty of an exercise is defined as the expected number of word errors the student would make in translating it. 
\end{itemize}
Given the above setting, we formalize our task as:
\begin{equation}
    \begin{aligned}
        e_{n+1} = \mathop{\arg\max}\limits_{e} P(e|{\bf s_{n+1}}, d_{n+1}, C_{n+1}),
    \end{aligned}
\end{equation}
where $e_{n+1}$ satisfies the following constraints:
\begin{gather}
    \forall c \in C_{n+1}: \exists i, {e_{n+1}}_{i: i+|c|} = c, \label{Eq.3} \\ 
    d_{n+1} = \sum_{w \in e_{n+1}} (1-{\bf s_{n+1}}[w]), \label{Eq.4}
\end{gather}
corresponding to \emph{word constraint} and \emph{difficulty constraint}, respectively. Here,
${\bf s_{n+1}}[w]$ represents the correct probability of translating word $w$; therefore, the sum of $\{1-{\bf s}[w]|, w \in e\}$ is the expected number of errors in translating $e$, which can be seen as a measure of the difficulty of $e$. 

Our task is distinct from previous CTG works in two aspects: 1) our control is \emph{dynamic}; student states acting as control are also learnable; 2) there is a strong dependency among control signals (Eqs. \ref{Eq.3} and \ref{Eq.4}), which is non-trivial to learn.
Note that in this work, we measure difficulty via student performance and only consider vocabulary knowledge in defining ${\bf s}$ for simplicity. Other definitions of sentence difficulty (e.g., definitions that incorporate other types of linguistic knowledge such as syntax) can be explored in future work.

\section{Methodology}\label{4}
\begin{figure}
\centering
\includegraphics[width=\columnwidth]{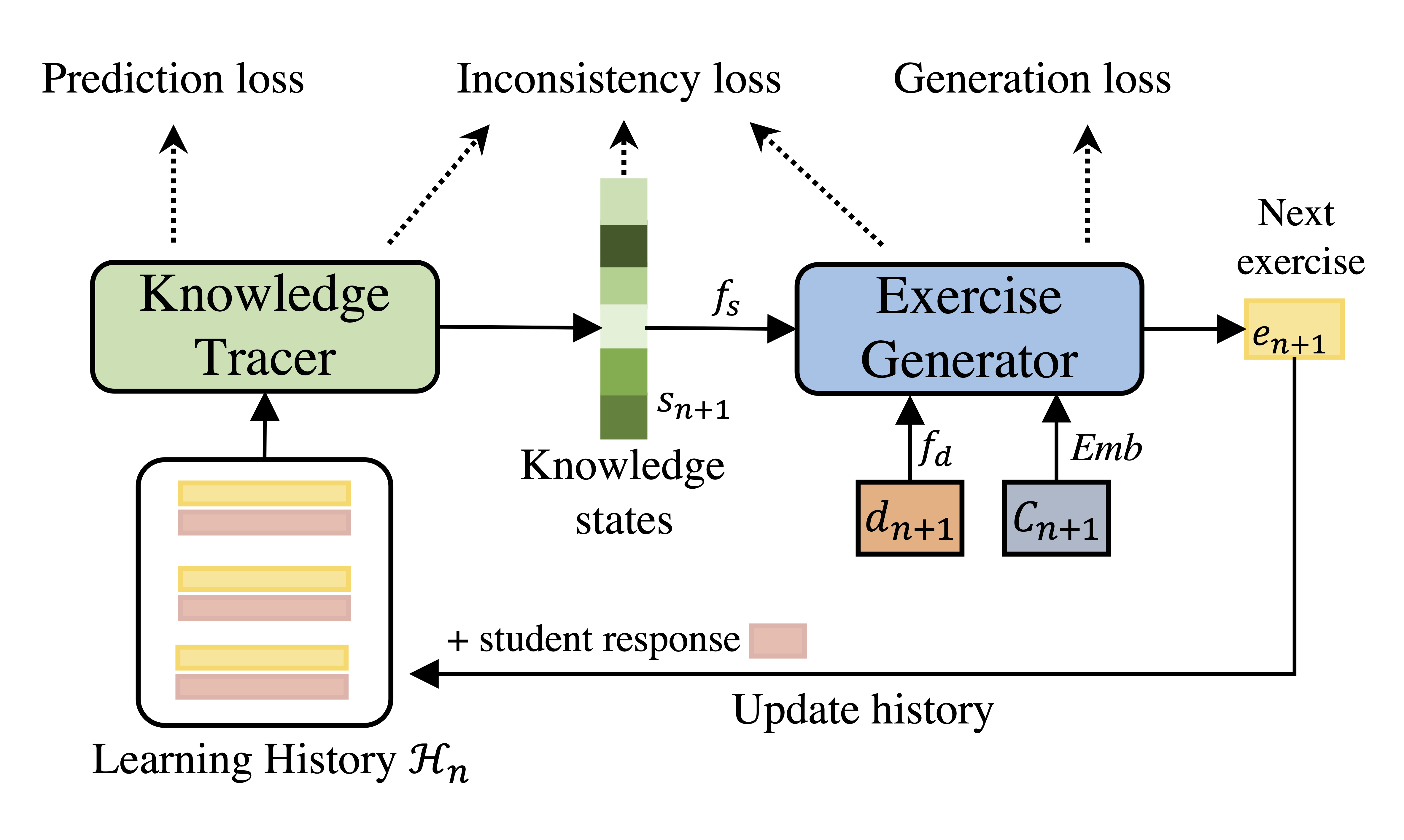}
\caption{The framework of our proposed model. We estimate a student's latest knowledge state ${\bf s_{n+1}}$ from the learning history $\mathcal{H}_{n}$, and then combine it with user-specified difficulty $d_{n+1}$ and knowledge components $C_{n+1}$ to generate the next exercise $e_{n+1}$. The two modules are jointly trained with an inconsistency loss to penalize their disagreement.}\label{figure-model}
\end{figure}

Our model is illustrated in Figure \ref{figure-model}.
We first employ a knowledge tracer $\mathcal{T}$ (\cref{4.1}) to estimate a student's time-varying knowledge states. Then, we build an LM-based exercise generator $\mathcal{G}$ (\cref{4.2}) to create exercises based on estimated states and specified difficulty and knowledge components (words). 
We jointly optimize the two modules with an inconsistency loss (\cref{4.3}) at training and apply a constrained decoding strategy (\cref{4.4}) at inference.
Finally, we discuss how our model can accommodate personalized learning recommendation algorithms on the fly (\cref{4.5}).
 
\subsection{Knowledge Tracing}\label{4.1}
The goal of our knowledge tracing model $\mathcal{T}$ is to estimate a student's latest knowledge state ${\bf s_{n+1}}$ given previous interactions $\mathcal{H}_{\leq n}$.
We adopt the deep knowledge tracing (DKT) model proposed by \citet{piech2015deep}.
We concatenate past exercises as a word sequence $ {\bf e}_{1:n} = \{w_{1,1},...,w_{n, |e_{n}|}\}$ and past responses as a label sequence ${\bf r}_{1:n} = \{r_{1,1},...,r_{n, |e_{n}|}\}$, where $w_{i,j}$ and $r_{i,j}$ represent the $jth$ word or label of the $ith$ exercise. 
Then we convert the two sequences into word embeddings $\vec{\bf e}_{1:n}$ and label embeddings $\vec{\bf r}_{1:n}$ and send them to an LSTM encoder to predict the next state ${\bf s_{n+1}}$:
\begin{gather}
    {\bf h_{n}} = {\rm LSTM}( \vec{\bf e}_{n} + \vec{\bf r}_{n}; {\bf h_{n-1}} ) \label{Eq.5}, \\
    {\bf s_{n+1}} = sigmoid({\rm W_{s}}*{\bf h_{n}} + {\rm b_{s}}) \label{Eq.6}.
\end{gather}
The model is trained to predict the binary word labels of the next exercise using the estimated knowledge state. 
The cross-entropy loss for a single student's history of $N$ interactions is computed as:
\begin{equation}
    \begin{aligned}
    \mathcal{L}_{ce} = \sum_{i=1}^{|N|} \sum_{j=1}^{|e_{i}|} {\rm CE}(r_{i,j}, {\bf s}_{i}[w_{i,j}]).
    \end{aligned}
\end{equation}

We adopt the regularization strategy proposed by \citet{yeung2018addressing} to stabilize training:
\begin{equation}
    \begin{aligned}
    \mathcal{L}_{r_{\{1,2\}}} = \sum_{n=2}^{N}\sum_{i=1}^{|\mathcal{V}|}|{\bf s_{n}}^{(i)}-{\bf s_{n-1}}^{(i)}|^{\{1,2\}}, \label{Eq.8}
    \end{aligned}
\end{equation}
where $\mathcal{L}_{r_1}$ ensures that only the states of relevant knowledge components are updated, and $\mathcal{L}_{r_2}$ penalizes the vibration. The final objective of $\mathcal{T}$ is $\mathcal{L}_{\mathcal{T}}=\mathcal{L}_{ce}+ \lambda_1*\mathcal{L}_{r_1} + \lambda_2*\mathcal{L}_{r_2}$ with $\lambda$ for balance.

\subsection{Controllable Exercise Generator}\label{4.2}
Our exercise generator $\mathcal{G}$ is fine-tuned from a pre-trained LM. Specifically, we generate an exercise $e$ based on a student's current knowledge state ${\bf s}$, target words $C$, and expected difficulty $d$ (we drop the interaction index to reduce clutter).
We parameterize the inputs as follows:
\begin{gather}
    {\bf x}=[f_{s}({\bf s}); f_{d}(d); Emb(c_{1},...,c_{|C|})],
\end{gather}
where knowledge state ${\bf s}$ and scalar difficulty $d$ are projected to control vectors via two feedforward layers $f_s$ and $f_d$, and $C$ are mapped to word embeddings. The training objective for generating a single exercise is defined as: 
\begin{gather}
    \mathcal{L}_{\mathcal{G}} = -\sum_{t}^{|e|} logP(w_{t}|w_{1}, ..., w_{t-1}, {\bf x}).
\end{gather}
During training, we sample a proportion of words from reference exercises as $C$ and calculate difficulty $d$ from ground-truth correctness labels, whereas states ${\bf s}$ are estimated by $\mathcal{T}$. 
At inference, $d$ and $C$ can be determined by instructors or the system, allowing automated and human intervention. \looseness=-1

\subsection{Joint Learning with Inconsistency Loss}\label{4.3}
We jointly optimize the knowledge tracer $\mathcal{T}$ and exercise generator $\mathcal{G}$ with an \emph{inconsistency loss} inspired by \citet{cui-hu-2021-topic-guided}, enabling the two modules to learn from each other.
Concretely, after generating an exercise $e$, we calculate its difficulty using input state ${\bf s}$ via Eq. \ref{Eq.4}, which should be as close to the input difficulty $d$ as possible:
 \begin{gather}
     \mathcal{L}_{inc} = |d - \sum_{w \in e}(1-{\bf s}[w])|.
 \end{gather}
 Since the second term is non-differentiable due to the $argmax$ operation involved in producing $e$, we replace it with "soft" tokens:
\begin{gather}
     \mathcal{L}_{inc} = |d - \sum_{t}^{|e|} (1-  {\bf p}_{t} \odot {\bf s})|,
\end{gather}
where ${\bf p}_{t}=softmax({\bf o}_{t}/ \tau)$ is the $t^{th}$ distribution normalized from its logits ${\bf o}_{t} \in \mathbb{R}^{|\mathcal{V}|}$ with a temperature parameter $\tau$, and $\odot$ represents dot product.

For the generator $\mathcal{G}$, this loss constrains the generation toward the target difficulty. For $\mathcal{T}$, the LM distributions $p_{\theta}$ provide similarity information between vocabulary words. This is analogous to the relationship of knowledge components, which has been shown helpful in knowledge tracing \citep{tong2020structure}. 
The final objective of our model is $\mathcal{L} = \mathcal{L}_{\mathcal{T}}+ \gamma_{1} \mathcal{L}_{\mathcal{G}}+ \gamma_{2} \mathcal{L}_{inc}$.

\subsection{Lexical Difficulty Constrained Decoding}\label{4.4}

We propose a beam search-based decoding algorithm to enforce the constraints introduced in \cref{3}.
At each step, we update the beam according to:
\begin{gather}
    Y_{t} \! = \hspace{-3mm} \mathop{{\rm argtopk}}\limits_{{\bf y}_{<t} \in Y_{t-1}, y_{t}\in \mathcal{V}} logP({\bf y}_{\leq t}|{\bf x})+\sum_{F_{i}\in \mathcal{F}}\alpha_{i}F_{i}({\bf y}_{\leq t}), \label{Eq.13}
\end{gather}
where $Y_{t}$ is the set of decoded hypotheses in step $t$ and $k$ is the beam size. 
The first term is the standard objective of beam search and the second term is a weighted combination of additional scoring functions in terms of the satisfaction of different constraints.
We formulate our constraints $\mathcal{F}$ in Eqs. \ref{Eq.3} and \ref{Eq.4} as:
\begin{gather}\nonumber
    F_c({\bf y}) = \sum_{c\in \mathcal{C}} I(c, {\bf y}), \hspace{1.5mm} \textit{and} \hspace{1.5mm} 
    F_d({\bf y}) = -|d - h({\bf y})|, \label{Eq.14}
\end{gather}
corresponding to the satisfaction of word constraint and difficulty constraint, respectively. $I(c, y)$ is a Boolean predicate indicating whether word $c$ is included in sequence ${\bf y}$ and $h({\bf y})$ calculates its difficulty via Eq. \ref{Eq.4}. 

Succinctly, the decoding algorithm works in three steps. 
First, we \textbf{expand} the current $k$ hypotheses to $k\times |\mathcal{V}|$ candidates. 
Then, we \textbf{prune} the search space by dropping candidates that are not in the top-$k_{F}$ list of any scoring functions $F$.
Finally, we \textbf{rescore} the pruned candidates based on the full objective (Eq. \ref{Eq.13}) and select the $k$-best ones to update the beam.

However, we found that greedily applying $F_d$ in the rescoring step would bias the decoder toward sequences with difficult words in the earlier steps.
Drawing inspiration from \citet{lu-etal-2022-neurologic}, we use lookahead heuristics that incorporate future estimates into the decoding process. 
Concretely, to score a subsequence ${\bf y}_{<t}$, we first greedily decode the next $l+1$ steps "soft" tokens (i.e., distributions): ${\bf \tilde{y}}_{t:t+l}$=$[{\bf p}_{t},..., {\bf p}_{t+l}]$.
Then, we combine the constraint satisfaction of decoded ${\bf y}_{<t}$ and the estimated future ${\bf \tilde{y}}_{t:t+l}$:   
\begin{gather*}
    \tilde{F}_c({\bf y}_{<t}) \! = \! \sum_{c\in \mathcal{C}} {\rm max}(I(c, {\bf y}_{<t}), \mathop{{\rm max}}\limits_{j \in [t, t+l] } P(y_{j}=c) ),   \\
    \tilde{F}_d({\bf y_{<t}}) = -|d - h({\bf y_{<t}}) - \sum_{j=t}^{t+l} 1- {\bf p}_{j} \odot {\bf s} |.
\end{gather*}
The procedure of our decoding algorithm is in Appendix \ref{A}.  
\subsection{Plug-and-Play Personalized Generation}\label{4.5}
Our model can be flexibly plugged into an existing personalized learning recommendation algorithm to automatically generate novel and customized exercises.
We showcase this functionality using the \textsc{ExpectiMax} curriculum planning strategy derived from DKT.
Given a student's current state ${\bf s_{n}}$, we can calculate the expected knowledge state after practicing a new exercise $e$ using our KT model $\mathcal{T}$:
\begin{gather}
    \tilde{{\bf s}}_{n+1} = \sum_{r \in \{0, 1\}^{|e|} } P(r) * \mathcal{T}({\bf s}_{n}, (e, r)),
\end{gather}
where $\mathcal{T}(\cdot)$ computes the updated knowledge state given a new interaction $(e,r)$. 
The probability of label sequence $r$ is computed from ${\bf s}_{n}$ assuming conditional independence $P(r) = \prod_{i=1}^{|e|} P(r_i)$, where $P(r_i)={\bf s}_{n}[e_{i}]$. 
\textsc{ExpectiMax} scores $e$ based on how well it can improve a student's average knowledge state, i.e., $F_{k}(e)=\overline{\tilde{{\bf s}}}_{n+1} - \overline{{\bf s}}_{n}$ \label{fk}, where $\overline{{\bf s}}$ denotes mean of the vector.
We incorporate $F_{k}$ into the decoding objective (Eq. \ref{Eq.13}) and call it \textsc{ExpectiMax-Gen}.

In principle, our model can accommodate different recommendation algorithms with different ranking functions $F_{k}$. 
The key benefit is that our model can \emph{generate} novel exercises, while retrieval-based systems can only \emph{select} exercises from an existing pool.

\begin{table}[]
\small
\centering
\begin{tabular}{@{}lcccc@{}}
\toprule
\multirow{2}{*}{\textbf{Model}} & \multicolumn{2}{c}{\textbf{Word-level}} & \multicolumn{2}{c}{\textbf{Exercise-level}} \\ \cmidrule(l){2-5} 
                                & \textbf{Seen}      & \textbf{Unseen}    & \textbf{Seen}        & \textbf{Unseen}      \\ \midrule
Ensemble                        & 73.41              & 70.58              & 65.55                & 64.93                \\
Standard DKT                             & 80.46              & 75.54              & 72.32                & 71.54                \\ \midrule
DKT$_{{\rm LM}, \tau=0.5}$    & 80.47            & 75.51               & {72.39}                & 71.47                \\
DKT$_{{\rm LM}, \tau=1.0}$     & 80.49             & 75.54              & {72.38}                & 71.49                \\
DKT$_{{\rm LM}, \tau=2.0}$         & \textbf{80.55}     & \textbf{75.69}     & \textbf{72.41}       & \textbf{71.74}       \\
DKT$_{{\rm LM}, \tau=3.0}$       & 80.54               & 75.48             & {72.33}                & 71.52               \\
DKT$_{{\rm LM}, \tau=5.0}$       & 80.31               & 75.46             & {72.28}                & 71.50               \\
\bottomrule
\end{tabular}
\caption{AUC ($\times$ 100) performance of knowledge tracing models on seen and unseen text examples. Exercise-level results are obtained by averaging word-level predictions.}
\label{table-kt-results}
\end{table}

\section{Experimental Results and Analysis}\label{6}
We experiment on the English track of Duolingo Second Language Acquisition Modeling (SLAM) dataset \citep{settles-etal-2018-second}, which contains about 1 million interactions of 2.6k learners over the first 30 days of learning a second language. For each student, we use the first 80\% of interactions for training, and the subsequent and the last 10\% for validation and testing, respectively.
Details of the dataset and experimental setup are in Appendix \ref{A2}.

We first evaluate the ability of the KT model to estimate student knowledge states in \cref{6.1}.
Then, we analyze the effectiveness of the exercise generator in \cref{6.2}.
Lastly, we showcase the superiority of our model in two educational scenarios with simulation experiments in \cref{6.3}. 

\begin{table*}[ht]
\small
\centering
\begin{tabular}{@{}lccccccccc@{}}
\toprule
\multirow{2}{*}{\textbf{Models}} & \multicolumn{2}{c}{\textbf{BLEU} $\uparrow$} & \multicolumn{2}{c}{\textbf{METEOR} $\uparrow$} & \multicolumn{2}{c}{\textbf{KC-Coverage (\%)} $\uparrow$} & \multicolumn{2}{c}{\textbf{D-MAE} $\downarrow$} & \multirow{2}{*}{\textbf{\begin{tabular}[c]{@{}c@{}} Invalid (\%) $\mathbf{\downarrow}$ \\ \end{tabular}}} \\ \cmidrule(lr){2-9}
                                 & \textbf{Seen}   & \textbf{Unseen} & \textbf{Seen}    & \textbf{Unseen}  & \textbf{Seen}         & \textbf{Unseen}       & \textbf{Seen}   & \textbf{Unseen}  &                                                                                      \\ \midrule
EG$_{\mathcal{H}}$                             & 9.23            & \textless{}0.01 & 18.79            & 6.05             & 14.26                 & 2.49                  & 0.396           & 1.500            & \textbf{0.071}                                                                       \\
AQG$_{\mathcal{H}+d}$                       & 10.28           & \textless{}0.01 & 20.15            & 7.16             & 15.84                 & 2.95                  & 0.463               & 0.985                & 1.674                                                                                \\
EG$_{C}$                             & 18.41           & 5.21            & 45.36            & 36.14            & \textbf{99.77}        & 90.63                 & 0.367           & 0.837            & 0.301                                                                                \\
EG$_{C+d}$                            & 11.84           & 15.94           & 40.89            & 42.10            & 96.23                 & 91.62                 & 0.564           & 0.679            & 0.385                                                                                \\ \midrule
APEG$_{{\bf s}+C+d}$                             & \textbf{22.47}  & \textbf{34.60}  & \textbf{56.15}   & \textbf{44.01}   & 99.61                 & \textbf{95.71}        & \textbf{0.246}  & \textbf{0.604}   & 0.283                                                                                \\
- joint learning                 & 22.01           & 33.15           & 55.80            & 42.85            & 99.63                 & 94.08                 & 0.251           & 0.619            & 0.281                                                                                \\
- constrained decoding           & 21.58           & 32.06           & 55.43            & 40.49            & 99.59                 & 94.77                 & 0.263           & 0.681            & 0.277                                                                                \\
\midrule
Upper bound                      & 53.65           & 41.24           & 74.97            & 52.10            & 99.75                 & 95.96                 & 0.060           & 0.302            & 0.233                                                                                \\ \bottomrule
\end{tabular}
\caption{Results of exercise generation.
APEG is our proposed model, and AQG is an adaptively difficulty-controlled question generation model proposed by \citet{srivastava-goodman-2021-question}.
The subscripts represent whether historical interactions ($\mathcal{H})$, target words ($C$), difficulty ($d$), and student state (${\bf s}$) are used to generate exercises.}
\label{table-eg-results}
\end{table*}

\subsection{Knowledge Tracing Evaluation}\label{6.1}
We use the standard \textbf{AUC (ROC)} as the metric of knowledge tracing in accordance with \citet{settles-etal-2018-second}. 
We denote our DKT model jointly trained with the LM-based exercise generator as DKT$_{\rm LM}$ and compare it with the following baselines: 
1) \underline{Ensemble} \citep{osika-etal-2018-second} which is one of the winning methods of the SLAM challenge that combines a RNN and a GBDT classifier.
We reimplement this model to use texts only as input and remove other side features, such as response time. 
We do this because we are interested in its performance in a \emph{general} setting where we do not assume the availability of diverse side information;
2) the \underline{standard DKT} \citep{piech2015deep} which is trained only with the KT loss $\mathcal{L}_{\mathcal{T}}$. 
We use it to verify whether jointly learning with an LM can help predict student language knowledge. 

We present the results in Table \ref{table-kt-results}, where we can see that DKT outperforms the Ensemble model when only text features are used, and our best model DKT$_{\rm LM, \tau=2}$ outperforms DKT on all metrics. 
We hypothesize the performance gain comes from the word similarity information entailed in the output distributions $p_{\theta}$ of the LM. 
This can be regarded as the relationship between knowledge components, which is demonstrated effective in knowledge tracing \citep{tong2020structure}.
To verify this, we tune the temperature $\tau$ which controls the sparsity of output distributions: 
$\tau \rightarrow 0$ produces a sparse distribution that is too assertive and provides little relationship information, while $\tau \rightarrow \infty $ produces a uniform distribution where all words are evenly related.
The results in the second section of Table \ref{table-kt-results} suggest that a medium $\tau$ improves the performance, while a small ($\tau$=1) or large ($\tau$=5) one is harmful, particularly for predicting unseen data.
The broader message from this observation is that the knowledge encoded in pre-trained LMs has the potential to improve knowledge tracing in the domain of language learning.
We also conduct an analysis of the influence of regularization terms Eq. \ref{Eq.8}, detailed in Appendix \ref{A3}.

\subsection{Exercise Generation Evaluation}\label{6.2}
The main results of exercise generation are presented in Table \ref{table-eg-results}, which are split according to whether the exercises are seen in the training set.
Evaluation metrics include reference-based \textbf{BLEU} \citep{papineni2002bleu} and \textbf{METEOR} \citep{banerjee2005meteor}, \textbf{KC-Coverage} which is the percentage of target knowledge components (words) that appear in the outputs, \textbf{D-MAE} which is the mean absolute error between the input difficulty and output difficulty, \textbf{Invalid} which is the percentage of exercises that have grammar errors detected using an automatic tool\footnote{\href{https://github.com/jxmorris12/language_tool_python}{https://github.com/jxmorris12/language\_tool\_python.}}. Since we generate exercises for language learning, we expect a valid exercise to be grammatically correct.
We analyze the performance from the following aspects.

\noindent \textbf{Lexical Controllability}. 
We first examine the lexical controllability of our model, which is crucial for generating personalized exercises for language learning.
We compare our model with two baselines:1) EG$_{\mathcal{H}}$ which generates the next exercise based on the student’s historical interactions; and 2) AGQ$_{\mathcal{H}+d}$\footnote{We obtain its results using the code released by the authors. Note that AQG is built on a different definition of difficulty. Thus, the D-MAE result might bias toward our model. We report this metric for reference only.} which generates the next exercise based on historical interactions and a target difficulty. 
The two baselines perform poorly on BLEU, METEOR, and KC-Coverage metrics, particularly for unseen data.
This indicates that they cannot predict the accurate content of the next exercise based on historical data or difficulty information, possibly because there is no strong connection within a sequence of exercises or such connection cannot be captured by an LM. 
We note that EG$_{\mathcal{H}}$ performs well on the validness metric. 
However, upon inspecting its results, we found the model almost only copies exercises from history, with less than 0.02\% novel generations.
The same issue is observed in AQG$_{\mathcal{H}+d}$ where more than 90\% exercises are repetitive. We follow \citet{srivastava-goodman-2021-question} to improve its novelty using a repetition penalty during the generation, but this results in far more invalid exercises (1.7\%).
In comparison, our model achieves a better balance between generalization ability and fluency. 

\noindent \textbf{Effect of Student Modeling}.
To investigate whether student modeling helps exercise generation, we build two baselines without student knowledge states: 1) EG$_{C}$ which conditions generation on target KCs (words) only, and 2) EG$_{C+d}$ on both target words and difficulty. 
The former variant can be considered a keyword-to-text generation model, while the latter imposes additional difficulty control. 
Our full model APEG$_{{\bf s}+C+d}$ significantly outperforms both of them, which proves our aforementioned hypothesis that a student's dynamic knowledge states must be considered in generating adaptive and personalized exercises.
An interesting observation is that incorporating difficulty control improves the performance on unseen data, indicating the model to some degree learns generalizable difficulty information.
Nevertheless, our further analysis shows the model is not adaptive to students of different abilities, which will be discussed in \cref{6.3}.

\begin{table}[t]
\centering
\small
\begin{tabular}{@{}lccc@{}}
\toprule
              & \textbf{BLEU} $\uparrow$  & \textbf{Coverage (\%)} $\uparrow$ & \textbf{D-MAE} $\downarrow$ \\ \midrule
w/o lookahead & 20.46          & 99.18                  & 0.263          \\
w/ lookahead  & \textbf{21.20} & \textbf{99.30}         & \textbf{0.257} \\ \bottomrule
\end{tabular}
 \caption{Comparison of generation performance with and without lookahead on the validation set.}
\label{table:ablation}
\end{table}

\begin{figure}
    \centering
    \includegraphics[width = 1.1\columnwidth, trim={5cm 0cm 0cm 5cm}, clip]{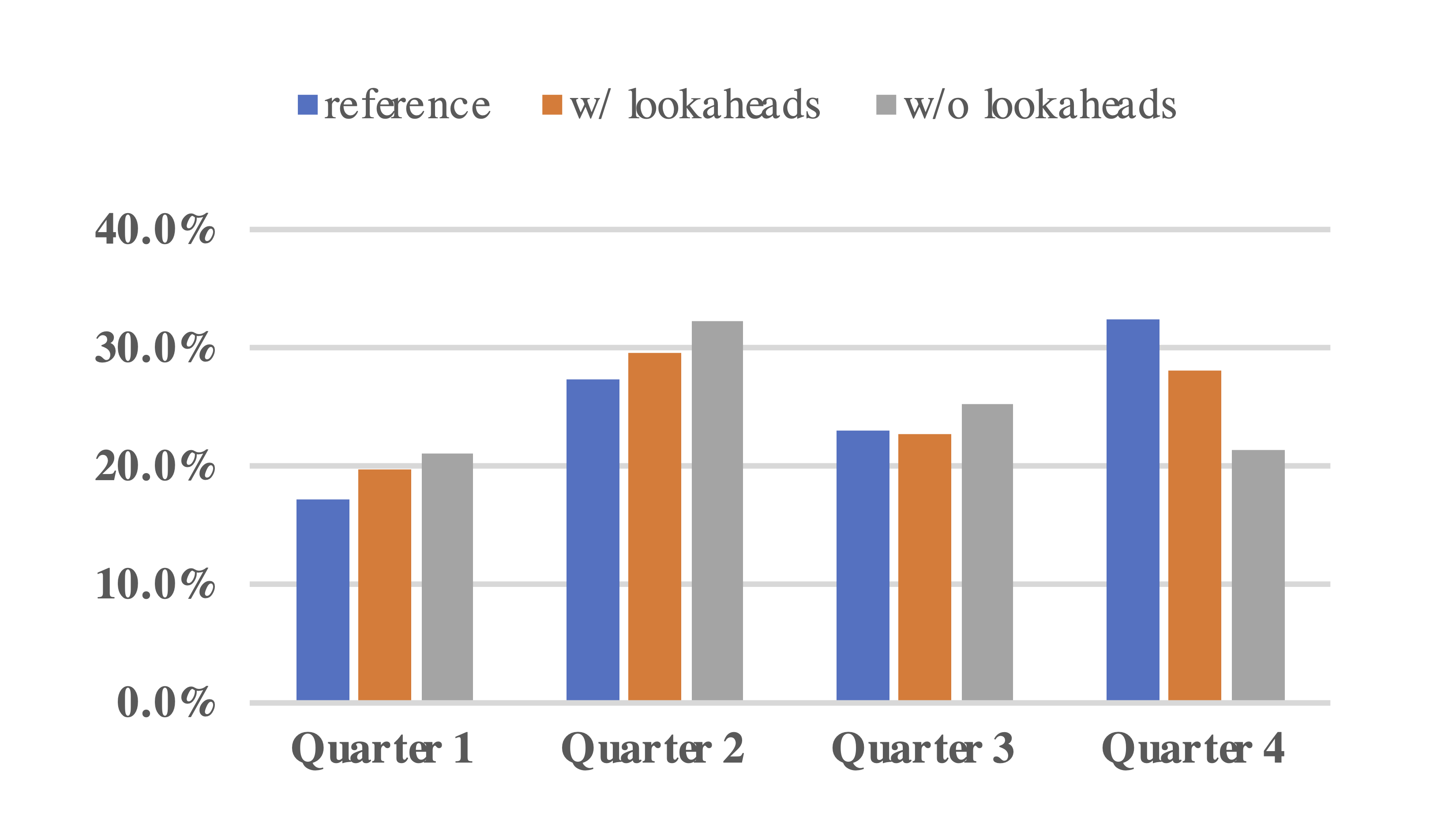}
    \caption{Distributions of accumulated word difficulty in four equally sized segments of 2000 sampled exercise sentences.}
    \label{fig:difficulty-dist}
\end{figure}

\noindent \textbf{Ablation Study}.
The key challenge of our task is to learn the dependency between student knowledge, vocabulary, and exercise difficulty (Eqs. \ref{Eq.3} and \ref{Eq.4}).
To understand which parts of our model contribute to this goal, we build two ablated variants by removing the joint learning strategy (\cref{4.3}) and the constrained decoding algorithm (\cref{4.4}), respectively. 
As shown in the second section of Table \ref{table-eg-results}, the search-based method is slightly better than the learning-based method, while combining them leads to the best performance. 

We further explore the effect of the lookahead strategy on difficulty constraints. 
Table \ref{table:ablation} presents the ablation results on the validation set, where we can see lookahead strategy improves both generation quality and controllability.
To understand how it works, we measure the distribution of difficulty in different regions of exercise sentences. 
Such distribution is computed as the accumulated word difficulty in four equally sized segments of 2000 sampled sentences.
As shown in Figure \ref{fig:difficulty-dist}, the difficult words of reference exercises are largely concentrated in the $2^{nd}$ and $4^{th}$ quarter. 
Our decoding algorithm with lookahead produces a similar result, while removing lookahead would bias the distribution toward $2^{nd}$ and $3^{rd}$ quarter. 
This confirms our assumption that naively applying $F_{d}$ would greedily select difficult words in the early steps, which is not the distribution of reference exercises. 
Our decoding algorithm avoids this issue by estimating the future and therefore achieves better results.

\noindent \textbf{Upper Bound Analysis}.
When we train our model, we use ground-truth difficulty $d$ and target words $C$ obtained from references; however, the student states ${\bf s}$ are estimated from the KT model.
We conduct an upper bound analysis to understand the influence of the accuracy of ${\bf s}$ on the generation performance.
Since a student's actual mastery of every vocabulary word is not available, we choose to replace the ground-truth difficulty levels $d$ with those estimated from ${\bf s}$.
As shown in the last section of Table 2, all metrics are considerably boosted when the inconsistency between states ${\bf s}$ and difficulty $d$ is eliminated.
This again proves the effect of incorporating student states and explains how such information comes to play: the knowledge states explicitly convey the dynamics between control signals $d$, $\mathcal{C}$, and target exercises $e$, which is non-trivial to learn by the model itself.

\begin{table}[t]
\footnotesize
\centering
\setlength\tabcolsep{2pt}
\begin{tabular}{llll}
\toprule
\textbf{${\bf d_{in}}$} & \textbf{Target words} & \textbf{Generated exercises} & \textbf{${\bf d_{out}}$} \\ \hline
\multicolumn{4}{c}{Avg. knowledge state $\overline{{\bf s}}$ = 0.32}                    \\ \hline
1.0        & \{men\}       & \adjustbox{margin=1.5pt,bgcolor=red!9.00}{Fifteen} \adjustbox{margin=1.5pt,bgcolor=red!50.38}{\underline{men}}.                          & 1.25           \\
2.0        & \{study\}       & \adjustbox{margin=1.5pt,bgcolor=red!8.57}{I} \adjustbox{margin=1.5pt,bgcolor=red!60.58}{\underline{study}} \adjustbox{margin=1.5pt,bgcolor=red!35.38}{English}.                           & 2.18           \\
3.0        & \{airport\}       &  \adjustbox{margin=1.5pt,bgcolor=red!28.38}{Where} \adjustbox{margin=1.5pt,bgcolor=red!25.38}{is} \adjustbox{margin=1.5pt,bgcolor=red!11.38}{the} \adjustbox{margin=1.5pt,bgcolor=red!42.38}{\underline{airport}}?                          & 2.73            \\
\hline
\multicolumn{4}{c}{Avg. knowledge state $\overline{{\bf s}}$ = 0.65}                    \\ \hline
1.0        & \{profile\}       & \adjustbox{margin=1.5pt,bgcolor=red!5.25}{He} \adjustbox{margin=1.5pt,bgcolor=red!5.15}{has} \adjustbox{margin=1.5pt,bgcolor=red!3.50}{a} \adjustbox{margin=1.5pt,bgcolor=red!8.15}{famous} \adjustbox{margin=1.5pt,bgcolor=red!45}{\underline{profile}}.                           & 0.94           \\
2.0        & \{white, bitter\}       & \adjustbox{margin=1.5pt,bgcolor=red!5.38}{The} \adjustbox{margin=1.5pt,bgcolor=red!7.38}{\underline{white}} \adjustbox{margin=1.5pt,bgcolor=red!55.38}{mushroom} \adjustbox{margin=1.5pt,bgcolor=red!6.98}{is} \adjustbox{margin=1.5pt,bgcolor=red!37.38}{\underline{bitter}}.                           & 1.75           \\
3.0        & \{hit, nail\}       & \adjustbox{margin=1.5pt,bgcolor=red!8.38}{She} \adjustbox{margin=1.5pt,bgcolor=red!39.38}{\underline{hit}} \adjustbox{margin=1.5pt,bgcolor=red!7.38}{the} \adjustbox{margin=1.5pt,bgcolor=red!46.38}{\underline{nail}} \adjustbox{margin=1.5pt,bgcolor=red!5.38}{on} \adjustbox{margin=1.5pt,bgcolor=red!4.38}{the} \adjustbox{margin=1.5pt,bgcolor=red!33.38}{head}.                          & 2.89           \\
\bottomrule
\end{tabular}
\caption{
Examples of exercises based on different controls. 
$d_{in}$ is the input difficulty while $d_{out}$ is the output difficulty estimated by our knowledge tracing model. 
The degree of \adjustbox{margin=1.5pt,bgcolor=red!30}{highlight} represents a student's mastery of vocabulary words (the darker the harder). 
}
\label{table-cases}
\end{table}

\noindent \textbf{Case Study}.
We provide a few cases in Table \ref{table-cases}.
We can see our model can dynamically adjust the exercise content according to specified words, target difficulty, as well as students' different mastery states of the vocabulary.
The exercises generated for advanced students (avg. state = 0.65) are generally more difficult than for poor students (avg. state = 0.32) under the same input difficulty. 
 
\subsection{Educational Applications}\label{6.3} 
In this subsection, we showcase the potential applications of our model in two educational scenarios with simulation experiments.

\begin{figure*}[t]
\centering
\includegraphics[width=2.05\columnwidth]{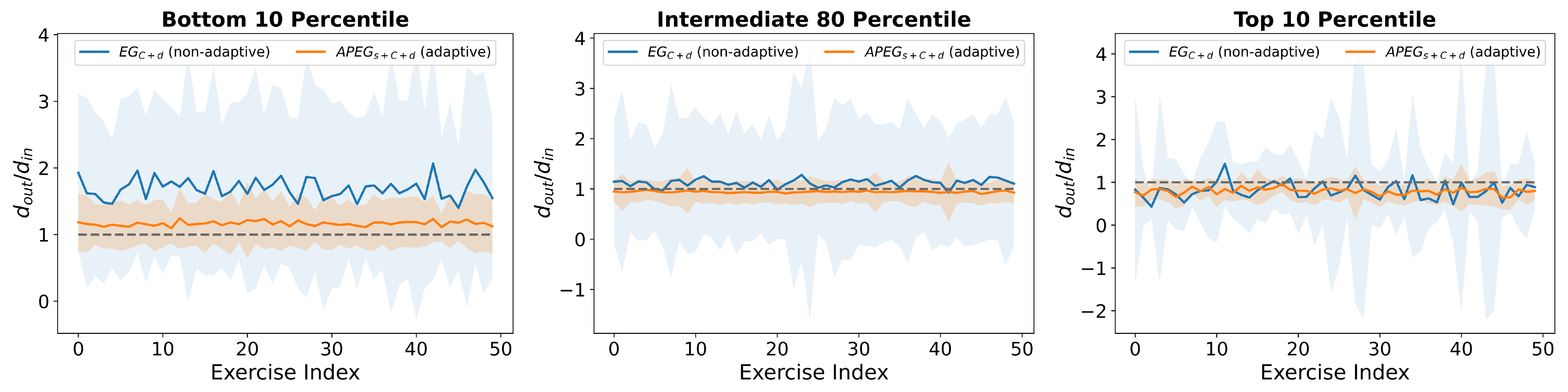}
\setlength{\abovecaptionskip}{0.55cm}
\caption{
Generating 50 additional exercises of specified difficulty levels for different student groups using APEG$_{s+C+d}$ (adaptive) and non-adaptive EG$_{C+d}$ models. 
The Y-axis is the ratio of output difficulty $d_{out}$ to input difficulty $d_{in}$; the closer to 1 (dotted line) the better. Solid lines are averaged results of group students at each step, and shadows represent standard deviations.
}
\label{figure-calibration}
\end{figure*}

\subsubsection{Adaptive Difficulty Calibration}\label{5.3.1}
A crucial requirement for adaptive learning systems is to dynamically adjust the difficulty of learning items to match each student's learning progress \citep{becker2018horizon}.
However, previous difficulty-controlled question generation approaches are mainly based on inherent problem difficulty, independent of individual abilities \citep{susanti2017controlling, kumar2019difficulty}.
Ideally, our model can achieve this goal by learning the dependency between difficulty and student knowledge states.
To verify this, we generate 50 additional exercises of specified difficulties for each student after their existing interactions.
At each step, we construct input by sampling a target word from the vocabulary and a difficulty level from a uniform distribution $[1, 3]$.
We compare our full model APEG$_{s+C+d}$ with its variant EG$_{C+d}$ which achieves the best difficulty controllability for unseen data. 
This baseline can be considered a vanilla non-adaptive difficulty-controlled exercise generation model.

In this simulation, we are interested in whether the difficulty controllability of our model can adapt to students of various knowledge levels.
To this end, we rank students based on their average knowledge states $\overline{{\bf s}}$ and split the result accordingly. 
As shown in Figure \ref{figure-calibration}, the difficulty controllability of the baseline is not reliable across different groups.
In particular, it tends to generate harder (up to $2 \times d_{in}$) exercises for the bottom 10 percentile students but easier (up to $\frac{1}{2} \times d_{in}$) ones for the top 10 percentile students, although it performs well for the intermediate 80 percentile students. 
In comparison, our adaptive model is also slightly biased toward the intermediate group but much more consistent than the baseline, with less than 20\% fluctuations on average.
Besides, we can see from the shadows that the baseline experiences huge variances at each step, indicating it is not adaptive to different knowledge states, even though the students within a group are at a similar level. 

\begin{figure}[t]
\centering
\includegraphics[width=\columnwidth]{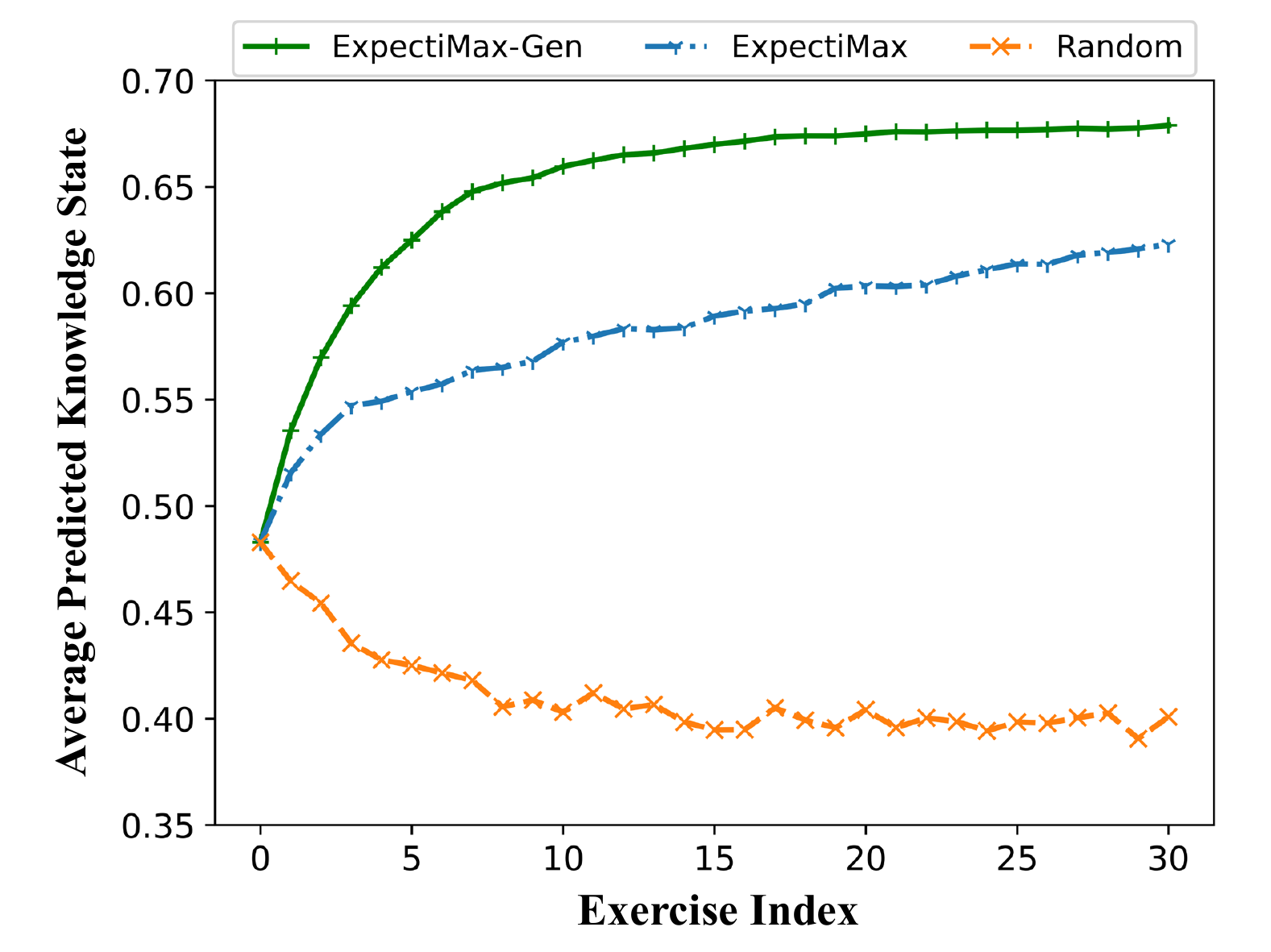}
\caption{Simulation results over 30 exercises.
The X-axis is the number of exercises, and the Y-axis is students' average predicted knowledge state $\overline{\tilde{{\bf s}}}$ indicating a student's overall mastery of the vocabulary.
}
\label{figure-simulation}
\end{figure}
\subsubsection{Improving Learning Efficiency} \label{5.3.2} 
We now examine whether our model can be used to improve student learning efficiency by personalizing exercise sequences. 
To this end, we customize 30 continuous exercises for 50 sampled students using our proposed \textsc{ExpectiMax-Gen} (\cref{4.5}) and the original \textsc{ExpectiMax}.
Both of them aim to maximize the expected knowledge state of the next step $\overline{\tilde{{\bf s}}}_{n+1}$.
For the former, at each step, we first find the best single word that can maximize $\overline{\tilde{{\bf s}}}_{n+1}$ and then generate the next exercise based on the selected word and a fixed difficulty of 1.
For the latter, we directly select the best exercise from the pool.
We update students' knowledge states after each practice and repeat this process until we collect 30 exercises. 
We compare the change in $\overline{\tilde{{\bf s}}}$ to measure which strategy is more efficient in improving students' knowledge.

The simulation results are shown in Figure \ref{figure-simulation}.
We also include a randomly selected exercise sequence as a lower bound, which turns out to harm student learning most of the time.
The decrease in knowledge state is possibly caused by overly difficult exercises which would lead to wrong answers and reduce the predicted probability.
Under the same practice opportunities, exercises generated by \textsc{ExpectiMax-Gen} lead to faster knowledge growth than those selected by \textsc{ExpectiMax}.
Upon further inspection, we found about 70\% of them are unseen in the corpus.
This explains the efficiency of \textsc{ExpectiMax-Gen} as it can create novel exercises targeting individual needs on the fly while \textsc{ExpectiMax} is limited by the pool. 
\looseness=-1

\subsubsection{Qualitative Discussions on Simulation} \label{5.3.3}
Our simulations are based on the DKT model. We note that some previous studies have observed inconsistencies between DKT behaviors and the human learning process \citep{shen2021process}. Thus, we adopt a simple regularization approach (Eqs. \ref{Eq.5} and \ref{Eq.6}) to alleviate such inconsistencies \citep{yeung2018addressing}, which we found can reduce the variance of simulation results and improve KT performance (Appendix \ref{A3}).

A popular argument regarding the relationship between the difficulty of learning content and student outcomes is that the level of difficulty should be set just above the learner's current knowledge, i.e., $d \approx 0.5$ \citep{settles2016trainable,gallego2018measuring}.
During the simulations, we found \textsc{ExpectiMax} does not follow this heuristic but tends to generate relatively easy exercises ($d<0.3$ mostly) repeatedly using certain words, consistent with the finding in \citet{tschiatschek2022equity}. 
One possible reason is that easier exercises are more likely to produce correct answers, which in turn increases the averaged predicted probability of DKT (i.e., estimated knowledge state).

Nevertheless, the above observations do not influence our conclusion as the superiority of our model comes from its ability to adapt to students' knowledge (\cref{5.3.1}) and generate customized exercises targeting individual needs (\cref{5.3.2}), independent of the simulation policy.  

\section{Conclusion}
We propose an adaptive and personalized exercise generation model combining recent advances in knowledge tracing and controllable generation using pre-trained LMs.
Our approach works by learning the dynamics between exercise difficulty and student vocabulary knowledge in the domain of language learning. 
Experimental results on real-world language learning data from Duolingo demonstrate that our model can generate adaptive and personalized exercises needed in an Educational setting. 
We further showcase our model's applicability in Education with simulation studies.

\section*{Ethics Statement}
The learner data used in this study are anonymized by \citet{settles-etal-2018-second} and, to the best of our knowledge, do not contain sensitive information. We foresee no further ethical or privacy concerns with the work.

\section*{Limitations}
We state the limitations of this work from the following aspects.
First, we make an initial assumption about the dynamics between exercise difficulty, vocabulary, and student knowledge. While we believe our assumption is sensible in the domain of language learning, we acknowledge that we make some simplifications for the ease of modeling. For example, we measure difficulty using individual performance, whereas a better way could be combining it with inherent problem difficulty, e.g., text complexity. Besides, we only consider vocabulary mastery in defining student knowledge and predicting their performance. Exploring more dimensions of language knowledge (e.g., syntax) might lead to a finer-grained personalization.
Second, our model relies on student learning logs to estimate their real-time knowledge states. This model might face the cold start problem when dealing with insufficient history. 
Though it is beyond the scope of this study, techniques like computerized adaptive testing can be used to combat this problem.
Lastly, due to the lack of a real learning environment, we discuss the educational promise of our model with simulation experiments. In the future, a user study can be incorporated to validate our conclusions.

\bibliography{custom}

\begin{thebibliography}{48}
\expandafter\ifx\csname natexlab\endcsname\relax\def\natexlab#1{#1}\fi

\bibitem[{Abdelrahman and Wang(2019)}]{abdelrahman2019knowledge}
Ghodai Abdelrahman and Qing Wang. 2019.
\newblock Knowledge tracing with sequential key-value memory networks.
\newblock In \emph{Proceedings of the 42nd International ACM SIGIR Conference
  on Research and Development in Information Retrieval}, pages 175--184.

\bibitem[{Agarwal and Mannem(2011)}]{agarwal2011automatic}
Manish Agarwal and Prashanth Mannem. 2011.
\newblock Automatic gap-fill question generation from text books.
\newblock In \emph{Proceedings of the sixth workshop on innovative use of NLP
  for building educational applications}, pages 56--64.

\bibitem[{Bailey et~al.(2018)Bailey, Vaduganathan, Henry, Laverdiere, and
  Pugliese}]{bailey2018making}
Allison Bailey, Nithya Vaduganathan, Tyce Henry, Renee Laverdiere, and Lou
  Pugliese. 2018.
\newblock Making digital learning work: Success strategies from six leading
  universities and community colleges.
\newblock \emph{Boston: Massachusetts: Boston Consulting Group}.

\bibitem[{Banerjee and Lavie(2005)}]{banerjee2005meteor}
Satanjeev Banerjee and Alon Lavie. 2005.
\newblock Meteor: An automatic metric for mt evaluation with improved
  correlation with human judgments.
\newblock In \emph{Proceedings of the acl workshop on intrinsic and extrinsic
  evaluation measures for machine translation and/or summarization}, pages
  65--72.

\bibitem[{Becker et~al.(2018)Becker, Brown, Dahlstrom, Davis, DePaul, Diaz, and
  Pomerantz}]{becker2018horizon}
Samantha~Adams Becker, Malcolm Brown, Eden Dahlstrom, Annie Davis, Kristi
  DePaul, Veronica Diaz, and Jeffrey Pomerantz. 2018.
\newblock Horizon report 2018 higher education edition brought to you by
  educause.
\newblock Technical report, EDUCAUSE.

\bibitem[{Cen et~al.(2008)Cen, Koedinger, and Junker}]{cen2008comparing}
Hao Cen, Kenneth Koedinger, and Brian Junker. 2008.
\newblock Comparing two irt models for conjunctive skills.
\newblock In \emph{International Conference on Intelligent Tutoring Systems},
  pages 796--798. Springer.

\bibitem[{Corbett and Anderson(1994)}]{corbett1994knowledge}
Albert~T Corbett and John~R Anderson. 1994.
\newblock Knowledge tracing: Modeling the acquisition of procedural knowledge.
\newblock \emph{User modeling and user-adapted interaction}, 4(4):253--278.

\bibitem[{Cui and Hu(2021)}]{cui-hu-2021-topic-guided}
Peng Cui and Le~Hu. 2021.
\newblock \href {https://doi.org/10.18653/v1/2021.findings-emnlp.126}
  {Topic-guided abstractive multi-document summarization}.
\newblock In \emph{Findings of the Association for Computational Linguistics:
  EMNLP 2021}, pages 1463--1472, Punta Cana, Dominican Republic. Association
  for Computational Linguistics.

\bibitem[{Daines et~al.(2016)Daines, Troka, and Santiago}]{daines2016improving}
Jennifer~B Daines, Tonya Troka, and John~M Santiago. 2016.
\newblock Improving performance in trigonometry and pre-calculus by
  incorporating adaptive learning technology into blended models on campus.
\newblock In \emph{2016 ASEE Annual Conference \& Exposition}.

\bibitem[{Dathathri et~al.(2020)Dathathri, Madotto, Lan, Hung, Frank, Molino,
  Yosinski, and Liu}]{Dathathri2020Plug}
Sumanth Dathathri, Andrea Madotto, Janice Lan, Jane Hung, Eric Frank, Piero
  Molino, Jason Yosinski, and Rosanne Liu. 2020.
\newblock \href {https://openreview.net/forum?id=H1edEyBKDS} {Plug and play
  language models: A simple approach to controlled text generation}.
\newblock In \emph{International Conference on Learning Representations}.

\bibitem[{Ficler and Goldberg(2017)}]{ficler-goldberg-2017-controlling}
Jessica Ficler and Yoav Goldberg. 2017.
\newblock \href {https://doi.org/10.18653/v1/W17-4912} {Controlling linguistic
  style aspects in neural language generation}.
\newblock In \emph{Proceedings of the Workshop on Stylistic Variation}, pages
  94--104, Copenhagen, Denmark. Association for Computational Linguistics.

\bibitem[{Gallego-Dur{\'a}n et~al.(2018)Gallego-Dur{\'a}n, Molina-Carmona, and
  Llorens-Largo}]{gallego2018measuring}
Francisco~J Gallego-Dur{\'a}n, Rafael Molina-Carmona, and Fara{\'o}n
  Llorens-Largo. 2018.
\newblock Measuring the difficulty of activities for adaptive learning.
\newblock \emph{Universal access in the information society}, 17:335--348.

\bibitem[{Heck and Meurers(2022)}]{heck2022parametrizable}
Tanja Heck and Detmar Meurers. 2022.
\newblock Parametrizable exercise generation from authentic texts: Effectively
  targeting the language means on the curriculum.
\newblock In \emph{Proceedings of the 17th Workshop on Innovative Use of NLP
  for Building Educational Applications (BEA 2022)}, pages 154--166.

\bibitem[{Holthaus et~al.(2019)Holthaus, Pancar, and
  Bergamin}]{holthaus2019recommendation}
Matthias Holthaus, Tansu Pancar, and Per Bergamin. 2019.
\newblock Recommendation acceptance in a simple adaptive learning system.

\bibitem[{Holtzman et~al.(2018)Holtzman, Buys, Forbes, Bosselut, Golub, and
  Choi}]{holtzman-etal-2018-learning}
Ari Holtzman, Jan Buys, Maxwell Forbes, Antoine Bosselut, David Golub, and
  Yejin Choi. 2018.
\newblock \href {https://doi.org/10.18653/v1/P18-1152} {Learning to write with
  cooperative discriminators}.
\newblock In \emph{Proceedings of the 56th Annual Meeting of the Association
  for Computational Linguistics (Volume 1: Long Papers)}, pages 1638--1649,
  Melbourne, Australia. Association for Computational Linguistics.

\bibitem[{Huang et~al.(2022)Huang, Liu, Chen, Hu, Liu, and
  Luo}]{huang2022design}
Shuyan Huang, Qiongqiong Liu, Jiahao Chen, Xiangen Hu, Zitao Liu, and Weiqi
  Luo. 2022.
\newblock A design of a simple yet effective exercise recommendation system in
  k-12 online learning.
\newblock In \emph{International Conference on Artificial Intelligence in
  Education}, pages 208--212. Springer.

\bibitem[{Imhof et~al.(2020)Imhof, Bergamin, and
  McGarrity}]{imhof2020implementation}
Christof Imhof, Per Bergamin, and St{\'e}phanie McGarrity. 2020.
\newblock Implementation of adaptive learning systems: Current state and
  potential.
\newblock \emph{Online teaching and learning in higher education}, pages
  93--115.

\bibitem[{K{\"a}ser et~al.(2017)K{\"a}ser, Klingler, Schwing, and
  Gross}]{kaser2017dynamic}
Tanja K{\"a}ser, Severin Klingler, Alexander~G Schwing, and Markus Gross. 2017.
\newblock Dynamic bayesian networks for student modeling.
\newblock \emph{IEEE Transactions on Learning Technologies}, 10(4):450--462.

\bibitem[{Keskar et~al.(2019)Keskar, McCann, Varshney, Xiong, and
  Socher}]{keskar2019ctrl}
Nitish~Shirish Keskar, Bryan McCann, Lav~R Varshney, Caiming Xiong, and Richard
  Socher. 2019.
\newblock Ctrl: A conditional transformer language model for controllable
  generation.
\newblock \emph{arXiv preprint arXiv:1909.05858}.

\bibitem[{Kumar et~al.(2019)Kumar, Hua, Ramakrishnan, Qi, Gao, and
  Li}]{kumar2019difficulty}
Vishwajeet Kumar, Yuncheng Hua, Ganesh Ramakrishnan, Guilin Qi, Lianli Gao, and
  Yuan-Fang Li. 2019.
\newblock Difficulty-controllable multi-hop question generation from knowledge
  graphs.
\newblock In \emph{International Semantic Web Conference}, pages 382--398.
  Springer.

\bibitem[{Lewis et~al.(2020)Lewis, Liu, Goyal, Ghazvininejad, Mohamed, Levy,
  Stoyanov, and Zettlemoyer}]{lewis2020bart}
Mike Lewis, Yinhan Liu, Naman Goyal, Marjan Ghazvininejad, Abdelrahman Mohamed,
  Omer Levy, Veselin Stoyanov, and Luke Zettlemoyer. 2020.
\newblock Bart: Denoising sequence-to-sequence pre-training for natural
  language generation, translation, and comprehension.
\newblock In \emph{Proceedings of the 58th Annual Meeting of the Association
  for Computational Linguistics}, pages 7871--7880.

\bibitem[{Liu et~al.(2020)Liu, Xu, Jia, Ma, Wang, and
  Vosoughi}]{liu-etal-2020-data}
Ruibo Liu, Guangxuan Xu, Chenyan Jia, Weicheng Ma, Lili Wang, and Soroush
  Vosoughi. 2020.
\newblock \href {https://doi.org/10.18653/v1/2020.emnlp-main.726} {Data boost:
  Text data augmentation through reinforcement learning guided conditional
  generation}.
\newblock In \emph{Proceedings of the 2020 Conference on Empirical Methods in
  Natural Language Processing (EMNLP)}, pages 9031--9041, Online. Association
  for Computational Linguistics.

\bibitem[{Lu et~al.(2022)Lu, Welleck, West, Jiang, Kasai, Khashabi, Le~Bras,
  Qin, Yu, Zellers, Smith, and Choi}]{lu-etal-2022-neurologic}
Ximing Lu, Sean Welleck, Peter West, Liwei Jiang, Jungo Kasai, Daniel Khashabi,
  Ronan Le~Bras, Lianhui Qin, Youngjae Yu, Rowan Zellers, Noah~A. Smith, and
  Yejin Choi. 2022.
\newblock \href {https://doi.org/10.18653/v1/2022.naacl-main.57}
  {{N}euro{L}ogic a*esque decoding: Constrained text generation with lookahead
  heuristics}.
\newblock In \emph{Proceedings of the 2022 Conference of the North American
  Chapter of the Association for Computational Linguistics: Human Language
  Technologies}, pages 780--799, Seattle, United States. Association for
  Computational Linguistics.

\bibitem[{Osika et~al.(2018)Osika, Nilsson, Sydorchuk, Sahin, and
  Huss}]{osika-etal-2018-second}
Anton Osika, Susanna Nilsson, Andrii Sydorchuk, Faruk Sahin, and Anders Huss.
  2018.
\newblock \href {https://doi.org/10.18653/v1/W18-0525} {Second language
  acquisition modeling: An ensemble approach}.
\newblock In \emph{Proceedings of the Thirteenth Workshop on Innovative Use of
  {NLP} for Building Educational Applications}, pages 217--222, New Orleans,
  Louisiana. Association for Computational Linguistics.

\bibitem[{Pandey and Karypis(2019)}]{PandeyK19-0}
Shalini Pandey and George Karypis. 2019.
\newblock \href
  {https://drive.google.com/file/d/18d_X6AXkPMhiHFQ2POarstVbX_7oMdFM} {A self
  attentive model for knowledge tracing}.
\newblock In \emph{Proceedings of the 12th International Conference on
  Educational Data Mining, EDM 2019, Montréal, Canada, July 2-5, 2019}.
  International Educational Data Mining Society (IEDMS).

\bibitem[{Papineni et~al.(2002)Papineni, Roukos, Ward, and
  Zhu}]{papineni2002bleu}
Kishore Papineni, Salim Roukos, Todd Ward, and Wei-Jing Zhu. 2002.
\newblock Bleu: a method for automatic evaluation of machine translation.
\newblock In \emph{Proceedings of the 40th annual meeting of the Association
  for Computational Linguistics}, pages 311--318.

\bibitem[{Perez and Cuadros(2017)}]{perez2017multilingual}
Naiara Perez and Montse Cuadros. 2017.
\newblock Multilingual call framework for automatic language exercise
  generation from free text.
\newblock In \emph{Proceedings of the Software Demonstrations of the 15th
  Conference of the European Chapter of the Association for Computational
  Linguistics}, pages 49--52.

\bibitem[{Piech et~al.(2015)Piech, Bassen, Huang, Ganguli, Sahami, Guibas, and
  Sohl-Dickstein}]{piech2015deep}
Chris Piech, Jonathan Bassen, Jonathan Huang, Surya Ganguli, Mehran Sahami,
  Leonidas~J Guibas, and Jascha Sohl-Dickstein. 2015.
\newblock Deep knowledge tracing.
\newblock \emph{Advances in neural information processing systems}, 28.

\bibitem[{Polozov et~al.(2015)Polozov, O'Rourke, Smith, Zettlemoyer, Gulwani,
  and Popovi{\'c}}]{polozov2015personalized}
Oleksandr Polozov, Eleanor O'Rourke, Adam~M Smith, Luke Zettlemoyer, Sumit
  Gulwani, and Zoran Popovi{\'c}. 2015.
\newblock Personalized mathematical word problem generation.
\newblock In \emph{Twenty-Fourth International Joint Conference on Artificial
  Intelligence}.

\bibitem[{Settles et~al.(2018)Settles, Brust, Gustafson, Hagiwara, and
  Madnani}]{settles-etal-2018-second}
Burr Settles, Chris Brust, Erin Gustafson, Masato Hagiwara, and Nitin Madnani.
  2018.
\newblock \href {https://doi.org/10.18653/v1/W18-0506} {Second language
  acquisition modeling}.
\newblock In \emph{Proceedings of the Thirteenth Workshop on Innovative Use of
  {NLP} for Building Educational Applications}, pages 56--65, New Orleans,
  Louisiana. Association for Computational Linguistics.

\bibitem[{Settles and Meeder(2016)}]{settles2016trainable}
Burr Settles and Brendan Meeder. 2016.
\newblock A trainable spaced repetition model for language learning.
\newblock In \emph{Proceedings of the 54th annual meeting of the association
  for computational linguistics (volume 1: Long papers)}, pages 1848--1858.

\bibitem[{Shen et~al.(2021)Shen, Liu, Chen, Huang, Huang, Yin, Su, and
  Wang}]{shen2021process}
Shuanghong Shen, Qi~Liu, Enhong Chen, Zhenya Huang, Wei Huang, Yu~Yin, Yu~Su,
  and Shijin Wang. 2021.
\newblock \href {https://doi.org/10.1145/3447548.3467237} {Learning
  process-consistent knowledge tracing}.
\newblock In \emph{Proceedings of the 27th ACM SIGKDD Conference on Knowledge
  Discovery \&amp; Data Mining}, KDD '21, page 1452–1460, New York, NY, USA.
  Association for Computing Machinery.

\bibitem[{Shin et~al.(2021)Shin, Shim, Yu, Lee, Kim, and Choi}]{shin2021saint+}
Dongmin Shin, Yugeun Shim, Hangyeol Yu, Seewoo Lee, Byungsoo Kim, and Youngduck
  Choi. 2021.
\newblock Saint+: Integrating temporal features for ednet correctness
  prediction.
\newblock In \emph{LAK21: 11th International Learning Analytics and Knowledge
  Conference}, pages 490--496.

\bibitem[{Srivastava and Goodman(2021)}]{srivastava-goodman-2021-question}
Megha Srivastava and Noah Goodman. 2021.
\newblock \href {https://doi.org/10.18653/v1/2021.acl-short.88} {Question
  generation for adaptive education}.
\newblock In \emph{Proceedings of the 59th Annual Meeting of the Association
  for Computational Linguistics and the 11th International Joint Conference on
  Natural Language Processing (Volume 2: Short Papers)}, pages 692--701,
  Online. Association for Computational Linguistics.

\bibitem[{Susanti et~al.(2017)Susanti, Tokunaga, Nishikawa, and
  Obari}]{susanti2017controlling}
Yuni Susanti, Takenobu Tokunaga, Hitoshi Nishikawa, and Hiroyuki Obari. 2017.
\newblock Controlling item difficulty for automatic vocabulary question
  generation.
\newblock \emph{Research and practice in technology enhanced learning},
  12(1):1--16.

\bibitem[{Tong et~al.(2020)Tong, Liu, Huang, Hunag, Chen, Liu, Ma, and
  Wang}]{tong2020structure}
Shiwei Tong, Qi~Liu, Wei Huang, Zhenya Hunag, Enhong Chen, Chuanren Liu,
  Haiping Ma, and Shijin Wang. 2020.
\newblock Structure-based knowledge tracing: an influence propagation view.
\newblock In \emph{2020 IEEE International Conference on Data Mining (ICDM)},
  pages 541--550. IEEE.

\bibitem[{Tschiatschek et~al.(2022)Tschiatschek, Knobelsdorf, and
  Singla}]{tschiatschek2022equity}
Sebastian Tschiatschek, Maria Knobelsdorf, and Adish Singla. 2022.
\newblock Equity and fairness of bayesian knowledge tracing.
\newblock \emph{arXiv preprint arXiv:2205.02333}.

\bibitem[{Vagale and Niedrite(2012)}]{vagale2012learner}
Vija Vagale and Laila Niedrite. 2012.
\newblock Learner model's utilization in the e-learning environments.
\newblock In \emph{DB\&Local Proceedings}, pages 162--174. Citeseer.

\bibitem[{Wang et~al.(2021)Wang, Lan, and Baraniuk}]{wang-etal-2021-math}
Zichao Wang, Andrew Lan, and Richard Baraniuk. 2021.
\newblock \href {https://doi.org/10.18653/v1/2021.emnlp-main.484} {Math word
  problem generation with mathematical consistency and problem context
  constraints}.
\newblock In \emph{Proceedings of the 2021 Conference on Empirical Methods in
  Natural Language Processing}, pages 5986--5999, Online and Punta Cana,
  Dominican Republic. Association for Computational Linguistics.

\bibitem[{Wolf et~al.(2020)Wolf, Debut, Sanh, Chaumond, Delangue, Moi, Cistac,
  Rault, Louf, Funtowicz et~al.}]{wolf2020transformers}
Thomas Wolf, Lysandre Debut, Victor Sanh, Julien Chaumond, Clement Delangue,
  Anthony Moi, Pierric Cistac, Tim Rault, R{\'e}mi Louf, Morgan Funtowicz,
  et~al. 2020.
\newblock Transformers: State-of-the-art natural language processing.
\newblock In \emph{Proceedings of the 2020 conference on empirical methods in
  natural language processing: system demonstrations}, pages 38--45.

\bibitem[{Wu et~al.(2020)Wu, Li, Tang, and Liang}]{wu2020exercise}
Zhengyang Wu, Ming Li, Yong Tang, and Qingyu Liang. 2020.
\newblock Exercise recommendation based on knowledge concept prediction.
\newblock \emph{Knowledge-Based Systems}, 210:106481.

\bibitem[{Yang and Klein(2021)}]{yang-klein-2021-fudge}
Kevin Yang and Dan Klein. 2021.
\newblock \href {https://doi.org/10.18653/v1/2021.naacl-main.276} {{FUDGE}:
  Controlled text generation with future discriminators}.
\newblock In \emph{Proceedings of the 2021 Conference of the North American
  Chapter of the Association for Computational Linguistics: Human Language
  Technologies}, pages 3511--3535, Online. Association for Computational
  Linguistics.

\bibitem[{Yarnall et~al.(2016)Yarnall, Means, and Wetzel}]{lessonsFrom}
Louise Yarnall, Barbara Means, and Tallie Wetzel. 2016.
\newblock \href {https://doi.org/10.13140/RG.2.2.36760.39688} {Lessons learned
  from early implementations of adaptive courseware}.

\bibitem[{Yeung and Yeung(2018)}]{yeung2018addressing}
Chun-Kit Yeung and Dit-Yan Yeung. 2018.
\newblock Addressing two problems in deep knowledge tracing via
  prediction-consistent regularization.
\newblock In \emph{Proceedings of the Fifth Annual ACM Conference on Learning
  at Scale}, pages 1--10.

\bibitem[{Yudelson et~al.(2013)Yudelson, Koedinger, and
  Gordon}]{yudelson2013individualized}
Michael~V Yudelson, Kenneth~R Koedinger, and Geoffrey~J Gordon. 2013.
\newblock Individualized bayesian knowledge tracing models.
\newblock In \emph{Artificial Intelligence in Education: 16th International
  Conference, AIED 2013, Memphis, TN, USA, July 9-13, 2013. Proceedings 16},
  pages 171--180. Springer.

\bibitem[{Zhao et~al.(2022)Zhao, Hou, Wang, Yu, Liu, and
  Ma}]{zhao2022educational}
Zhenjie Zhao, Yufang Hou, Dakuo Wang, Mo~Yu, Chengzhong Liu, and Xiaojuan Ma.
  2022.
\newblock Educational question generation of children storybooks via question
  type distribution learning and event-centric summarization.
\newblock In \emph{Proceedings of the 60th Annual Meeting of the Association
  for Computational Linguistics (Volume 1: Long Papers)}, pages 5073--5085.

\bibitem[{Zhou and Huang(2019)}]{zhou-huang-2019-towards}
Qingyu Zhou and Danqing Huang. 2019.
\newblock \href {https://doi.org/10.18653/v1/W19-8661} {Towards generating math
  word problems from equations and topics}.
\newblock In \emph{Proceedings of the 12th International Conference on Natural
  Language Generation}, pages 494--503, Tokyo, Japan. Association for
  Computational Linguistics.

\bibitem[{Ziegler et~al.(2019)Ziegler, Stiennon, Wu, Brown, Radford, Amodei,
  Christiano, and Irving}]{ziegler2019fine}
Daniel~M Ziegler, Nisan Stiennon, Jeffrey Wu, Tom~B Brown, Alec Radford, Dario
  Amodei, Paul Christiano, and Geoffrey Irving. 2019.
\newblock Fine-tuning language models from human preferences.
\newblock \emph{arXiv preprint arXiv:1909.08593}.

\end{thebibliography}
\bibliographystyle{acl_natbib}

\clearpage
\newpage

\appendix
\section{Decoding Algorithm}\label{A}
\begin{algorithm}[h]
\caption{Pseudo-code for our Lexical Difficulty Constrained Decoding}
\label{alg:DCQGFramework}
\small
\setstretch{1.2}
    \begin{algorithmic}[1]
    \Require Target words $\mathcal{C}$, difficulty $d$, a collection of score functions $\mathcal{F}$ and their weights $\alpha$, max step $T$, beam size $k$
    \Ensure  $k$ hypotheses $Y_{T}$ in the last step
        \State $Y_{0} \leftarrow \mathbf{InitBeam}()$ \Comment{\{<BOS>\}}
        \For{$t=1, t \leq T, t$++}
            \State  $ Y_{t} \leftarrow \varnothing$
            \State ${\rm Candidates} \leftarrow \mathbf{Generate}(Y_{t-1}, 1)$ \Comment{expand}
            \For{$F \in \mathcal{F}$} \Comment{prune candidates}
                \State $Y_{t} \leftarrow Y_{t} \cup \mathop{{\rm argtopk}}\limits_{{\bf y}_{\leq t} \in {\rm Candidates}} F({\bf y}_{\leq t}) $
            \EndFor

            \For{${\bf y}_{\leq t} \in Y_{t}$} \Comment{generate $l$-step lookaheads} 
                \State $\tilde{\bf y}_{t+1:t+l} = \mathbf{Generate}(y_{\leq t}, l)$ 
            \EndFor
            \State $Y_{t} \leftarrow  \mathop{{\rm argtopk}}\limits_{{\bf y}_{\leq t} \in Y_{t}} \sum_{F_{i} \in \mathcal{F}} \alpha_{i} F_{i}( {\bf y}_{\leq t} \circ \tilde{\bf y}_{t+1:t+l}) $   
        \EndFor
    \State \textbf{return} $Y_{T}$
    \end{algorithmic}
\end{algorithm}

\section{Experimental Setup} \label{A2}

\subsection{Dataset Details}
The statistics of our dataset are summarized in Table \ref{table-dataset}. 
Each interaction records a target sentence, per-token correctness labels of the student's response, and meta information such as user nationality and response time.
We group interactions by \texttt{user\_id} (anonymous) in temporal order to obtain per-student interaction sequences.
Refer to \citet{settles-etal-2018-second} for more descriptions of the dataset.
\begin{table}[h]
\centering
\small
\begin{tabular}{@{}lccc@{}}
\toprule
\multirow{2}{*}{\textbf{Statistics}} & \multicolumn{3}{c}{\textbf{Split}}            \\ \cmidrule(l){2-4} 
                                     & \textbf{Train} & \textbf{Dev} & \textbf{Test} \\ \midrule
\# of students                       & 2,593          & 2,593        & 2,593         \\
\# of interactions                   & 824,012        & 115,770      & 114,586       \\
\# of questions                      & 7,780           & 5,524             &  5,847             \\
\# of words (KCs)                         & 1,967               &  1,839            & 1,879              \\ \bottomrule
\end{tabular}
\caption{The statistics of SLAM English track.}
\label{table-dataset}
\end{table}

\subsection{Implementation Details}
We implement our models using the Transformers library \citep{wolf2020transformers}\footnote{\href{https://huggingface.co/docs/transformers/index}{https://huggingface.co/docs/transformers/index}}.
Our knowledge tracing model is a three-layer LSTM with a hidden size of 100. 
We train it for 10 epochs with the regularization weights $\lambda_{1}=0.5, \lambda_{2}=0.1$, selected on the validation set.
For the exercise generator, we fine-tune a pre-trained BART-base \citep{lewis2020bart} for up to 10 epochs. 
An early stop strategy is applied when the loss on the validation set does not decrease for three continuous epochs. 
We first train the DKT and exercise generator separately until both of them converge. Then, we jointly optimized the two models with hypearameters: $\gamma_{1}=1, \gamma_{2}=0.8, \tau=2$.
During generation, we set the beam size to 4. The weights $\alpha$ for word and difficulty constraints are set to 0.1 and 0.5 as the word constraint is easy to achieve in our experiments.
We use Nvidia Tesla A100 with 40 GB of GPU memory for training and inference.
On a single GPU, one training epoch of the exercise generator takes about 30 minutes, and that of DKT takes about 7 minutes when they are separately trained.
Joint training takes a longer time, about an hour for one epoch.
We report the average results over three runs. 

\section{Influence of Regularization in KT} \label{A3}
To inspect the influence of regularization terms (Eq. \ref{Eq.8}) on the KT performance, we conduct a grid search for $\lambda_{1}$ and $\lambda_{2}$ on the validation set.
As can be seen from Table \ref{table-reg1} and Table \ref{table-reg2}, $\mathcal{L}_{r_{1}}$ consistently improves exercise-level performance at the cost of sacrificing word-level performance, whereas $\mathcal{L}_{r_{2}}$ with a suitable weight ($\lambda_2=0.3$) can improve both in most cases. 
This suggests the students' knowledge states transit gradually over time. 
We choose $\lambda_{1}=0.5, \lambda_{2}=0.1$ for the best balance.

\begin{table}[h]
\centering
\begin{tabular}{@{}c|cccc@{}}
\diagbox{$\lambda_{1}$}{AUC}{$\lambda_{2}$}          & \textbf{0.0} & \textbf{0.1} & \textbf{0.3} & \textbf{0.5} \\ \hline
\textbf{0.0} & \adjustbox{margin=1.5pt,bgcolor=red!47}{79.51}        & \adjustbox{margin=1.5pt,bgcolor=red!45}{79.50}        & \adjustbox{margin=1.5pt,bgcolor=red!56.00}{79.57}        & \adjustbox{margin=1.5pt,bgcolor=red!35.00}{79.53}        \\
\textbf{0.1} & \adjustbox{margin=1.5pt,bgcolor=red!36.00}{79.44}        & \adjustbox{margin=1.5pt,bgcolor=red!38.00}{79.45}        & \adjustbox{margin=1.5pt,bgcolor=red!44.00}{79.49}        & \adjustbox{margin=1.5pt,bgcolor=red!48.00}{79.52}        \\
\textbf{0.3} & \adjustbox{margin=1.5pt,bgcolor=red!33.00}{79.42}        & \adjustbox{margin=1.5pt,bgcolor=red!30.00}{79.40}        & \adjustbox{margin=1.5pt,bgcolor=red!36.00}{79.44}        & \adjustbox{margin=1.5pt,bgcolor=red!24.00}{79.36}        \\
\textbf{0.5} & \adjustbox{margin=1.5pt,bgcolor=red!18}{79.32}        & \adjustbox{margin=1.5pt,bgcolor=red!35.00}{79.43}        & \adjustbox{margin=1.5pt,bgcolor=red!32.00}{79.41}        & \adjustbox{margin=1.5pt,bgcolor=red!15}{79.30}       
\end{tabular}
\caption{Validation results (AUC$\times$100) of word-level prediction under varying regularization weights.}
\label{table-reg1}
\end{table}

\begin{table}[h]
\centering
\begin{tabular}{@{}c|cccc@{}}
\diagbox{$\lambda_{1}$}{AUC}{$\lambda_{2}$}          & \textbf{0.0} & \textbf{0.1} & \textbf{0.3} & \textbf{0.5} \\ \hline
\textbf{0.0} & \adjustbox{margin=1.5pt,bgcolor=red!14.00}{70.89}  & \adjustbox{margin=1.5pt,bgcolor=red!23.00}{70.98}  & \adjustbox{margin=1.5pt,bgcolor=red!10.00}{70.85} & \adjustbox{margin=1.5pt,bgcolor=red!40.00}{71.15}  \\
\textbf{0.1} & \adjustbox{margin=1.5pt,bgcolor=red!29.00}{71.04}  & \adjustbox{margin=1.5pt,bgcolor=red!27.00}{71.02}  & \adjustbox{margin=1.5pt,bgcolor=red!31.00}{71.06} & \adjustbox{margin=1.5pt,bgcolor=red!48.00}{71.23}  \\
\textbf{0.3} & \adjustbox{margin=1.5pt,bgcolor=red!62.00}{71.41}  & \adjustbox{margin=1.5pt,bgcolor=red!56.00}{71.31}  & \adjustbox{margin=1.5pt,bgcolor=red!64.00}{71.43} & \adjustbox{margin=1.5pt,bgcolor=red!56.00}{71.31}  \\
\textbf{0.5} & \adjustbox{margin=1.5pt,bgcolor=red!62.00}{71.41}  & \adjustbox{margin=1.5pt,bgcolor=red!69.00}{71.48}  & \adjustbox{margin=1.5pt,bgcolor=red!66.00}{71.45} & \adjustbox{margin=1.5pt,bgcolor=red!66.00}{71.45}   \\ 
\end{tabular}
\caption{Validation results (AUC$\times$100) of exercise-level prediction under varying regularization weights.}
\label{table-reg2}
\end{table}

\end{document}